\documentclass[lettersize,journal]{IEEEtran}
\usepackage{amsmath,amsfonts}
\usepackage{algorithmic}
\usepackage{algorithm}
\usepackage{array}
\usepackage[caption=false,font=normalsize,labelfont=sf,textfont=sf]{subfig}
\usepackage{textcomp}
\usepackage{stfloats}
\usepackage{url}
\usepackage{verbatim}
\usepackage{graphicx}
\usepackage{cite}
\usepackage{xcolor}
\usepackage{multirow} 
\begin{document}

\title{Dual-Domain Homogeneous Fusion with Cross-Modal Mamba and Progressive Decoder for 3D Object Detection}

\author{{Xuzhong Hu, Zaipeng Duan, Pei An, Jun zhang, and Jie Ma}
\thanks{}
\thanks{}}

\markboth{Journal of \LaTeX\ Class Files,~Vol.~14, No.~8, August~2021}%
{Shell \MakeLowercase{\textit{et al.}}: A Sample Article Using IEEEtran.cls for IEEE Journals}


\maketitle

\begin{abstract}
Fusing LiDAR and image features in a homogeneous BEV domain has become popular for 3D object detection in autonomous driving. However, this paradigm is constrained by the excessive feature compression. While some works explore dense voxel fusion to enable better feature interaction, they face high computational costs and challenges in query generation. Additionally, feature misalignment in both domains results in suboptimal detection accuracy. To address these limitations, we propose a Dual-Domain Homogeneous Fusion network (DDHFusion), which leverages the complementarily of both BEV and voxel domains while mitigating their drawbacks. Specifically, we first transform image features into BEV and sparse voxel representations using lift-splat-shot and our proposed Semantic-Aware Feature Sampling (SAFS) module. The latter significantly reduces computational overhead by discarding unimportant voxels. Next, we introduce Homogeneous Voxel and BEV Fusion (HVF and HBF) networks for multi-modal fusion within respective domains. They are equipped with novel cross-modal Mamba blocks to resolve feature misalignment and enable comprehensive scene perception. The output voxel features are injected into the BEV space to compensate for the information loss caused by direct height compression. During query selection, the Progressive Query Generation (PQG) mechanism is implemented in the BEV domain to reduce false negatives caused by feature compression. Furthermore, we propose a Progressive Decoder (QD) that sequentially aggregates not only context-rich BEV features but also geometry-aware voxel features with deformable attention and the Multi-Modal Voxel Feature Mixing (MMVFM) block for precise classification and box regression. On the NuScenes dataset, DDHFusion achieves state-of-the-art performance, and extensive experiments demonstrate its superiority over other representative homogeneous fusion methods.

\end{abstract}


\section{Introduction}
\IEEEPARstart{A}{utonomous} driving has emerged as a prominent research frontier, advancing rapidly with the integration of deep learning models. Among its technologies, 3D object detection plays a pivotal role in enabling safe and efficient driving planning. In current autonomous systems, LiDAR and cameras serve as essential sensors for environmental perception. LiDAR utilizes laser echo technology to capture precise 3D spatial information in the form of point clouds, while cameras generate images rich in texture details by capturing visible light. By harnessing the complementary strengths of LiDAR and cameras, fusion-based object detectors have consistently outperformed their single-sensor counterparts in terms of accuracy and robustness. However, multi-sensor feature fusion is non-trivial due to the inherent heterogeneity between camera and LiDAR data. Many existing works \cite{vora_pointpainting_2020, zhao_sem-aug_2022-1, wang_pa3dnet_2023, xu_fusionpainting_2021, 2021ImprovingDA, Yang2024SACINetSC} simply enrich LiDAR points or voxels with image features by camera-to-LiDAR projection. However, this approach suffers from the loss of semantic information as only a small fraction of image features are preserved, leading to degraded detection performance in areas with sparse point distributions. Recent studies propose homogeneous fusion strategies \cite{liu_bevfusion_2023, liang_bevfusion_2022, Song2024GraphBEVTR, Yang2022DeepInteraction3O, yang_deepinteraction_2024, Li2022UnifyingVR} that actively transform image features into a shared space with LiDAR features. These approaches employ symmetric network architectures to exploit the combined advantages of multi-modal inputs, while simultaneously reducing the performance degradation that arises from single-sensor failures

\begin{figure}[tbp]
	\centering
	\includegraphics[width=0.5\textwidth]{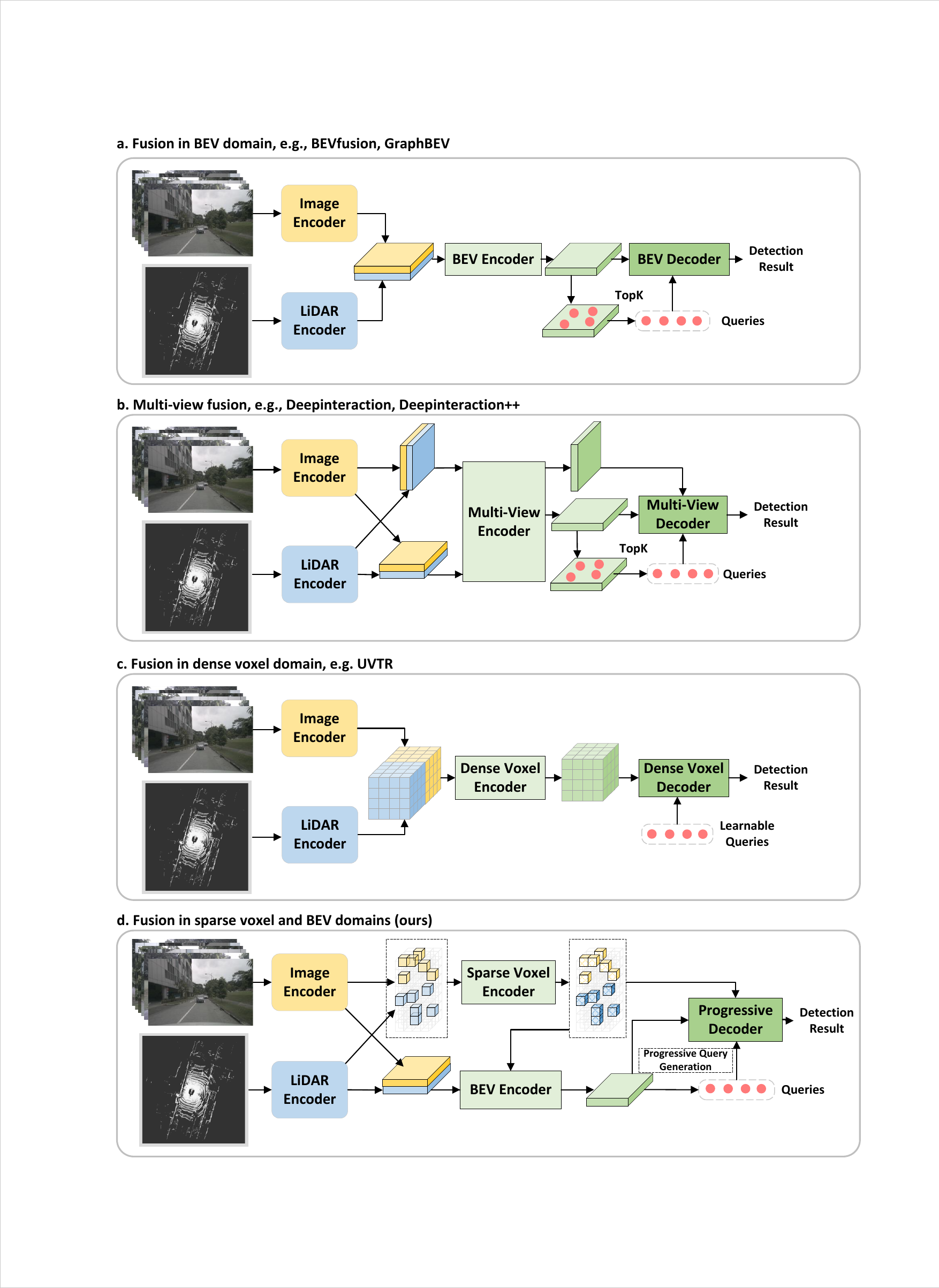} 
	\caption{Overview of homogeneous fusion methods. The view transformation in Fig.~\ref{fig:comp}(a) and Fig.~\ref{fig:comp}(b) causes the loss of modality details. The dense voxel representation in Fig.~\ref{fig:comp}(c) causes computational burden. Our method in Fig.~\ref{fig:comp}(d) combines the advantages of sparse voxel and BEV domains. } 
	\label{fig:comp}
\end{figure}

A popular paradigm \cite{liu_bevfusion_2023, liang_bevfusion_2022, Song2024GraphBEVTR}, as depicted in Fig.~\ref{fig:comp}(a), employs Lift-Splat-Shoot (LSS) to project image features onto a unified BEV plane, where they are concatenated with LiDAR features. Its framework is highly efficient as mature techniques from 2D detection tasks can be easily adopted for BEV feature fusion and query generation. For the latter, the simple center-based label assignment and Top-K selection strategies are widely applied. However, there remain two limitations: \textbf{1) Information Compression}: Collapsing geometric and texture information along the height dimension results in blurred features, which further leads to missed detections and reduced regression accuracy. As shown in Fig.~\ref{fig:comp}(b), some works \cite{Yang2022DeepInteraction3O, yang_deepinteraction_2024} introduce multi-view encoder-decoder structures to address this. However, the additional transformation of LiDAR features to the frontal view significantly disrupts 3D spatial relationships, still incurring feature ambiguity. \textbf{2) Feature Misalignment}: Inaccurate depth estimation in LSS between sensors may lead to spatial deviations of multi-modal BEV features. Extra alignment techniques are required before feeding the features into the BEV backbone.
 
As shown in Fig.~\ref{fig:comp}(c), fusion in a unified dense voxel space \cite{Li2022UnifyingVR} has been explored to and leverage higher-dimensional feature interactions while avoiding information compression. While theoretically superior to BEV-based fusion, it faces several challenges: \textbf{1) Computational Burden}: Fusing features represented as dense voxels is highly memory-intensive and computationally expensive, making it difficult to deploy in large-scale driving scenarios. \textbf{2) Query Generation Difficulty}: According to \cite{Deng2022VISTAB3}, compared to the BEV domain, the more severe foreground-background imbalance and higher dimensionality in the dense voxel domain complicate label assignment and input-dependent query generation \cite{Bai2022TransFusionRL}. \cite{Li2022UnifyingVR} adopts randomly initialized 3D learnable queries combined with local 3D deformable attention in decoders, which slows down training convergence and limits the model's ability to localize potential instances. \textbf{3) Feature Misalignment}: During feature transformation, the depth estimation, similar to that in LSS, also leads to feature misalignment. Worse still, computational constraints compel the network to rely on 3D convolutions for local fusion within limited receptive fields, which hinders feature alignment over larger regions.

In this work, we propose a novel detection framework—the Dual-Domain Homogeneous Fusion network (DDHFusion), which overcomes the limitations of BEV and voxel fusion domains while leveraging their respective strengths. As shown in Fig.~\ref{fig:comp}(d), the BEV domain is responsible for query generation, while the voxel domain provides geometry-aware features during feature encoding and decoding. For clarity, the proposed DDHFusion is divided into three steps: \textbf{1) Feature Transformation}: First, we transform image features into sparse voxel and BEV spaces using the semantic-aware feature sampling (SAFS) and LSS modules. Compared to \cite{Li2022UnifyingVR}, the sparse voxel representation significantly reduces computational overhead.  \textbf{2) Feature Encoding}: During encoding, these multi-representation image features interact with LiDAR ones through homogeneous voxel and BEV fusion (HVF and HBF) networks. Within these networks, we meticulously designs cross-modal voxel and BEV mamba blocks (CV-Mamba and CB-Mamba) to address feature misalignment and enable efficient global scene perception. \textbf{3) Feature Decoding}: For decoding, we propose a progressive query generation (PQG) module and a progressive decoder (PD). The former activates BEV features with the spatial relationship of high-response queries, generating potential hard queries in the second stage to mitigate false negatives caused by feature compression and small object sizes. The latter level up the conventional BEV decoder by integrating our proposed multi-modal voxel feature mixing (MMVFM) module which can aggregate fine-grained 3D structural information around targets for precise box regression.

As a summary, the main contributions are as follows:

\begin{enumerate}
	\item We present DDHFusion, a novel framework for 3D object detection that achieves state-of-the-art performance on the NuScenes dataset. It surpasses other homogeneous fusion methods in accuracy while maintaining efficient inference.
	\item To the best of our knowledge, we are the first to extend Mamba into the multi-modal 3D detection task. The proposed CV-Mamba and CB-Mamba blocks effectively facilitate feature fusion across global voxel and BEV domains.
	\item We propose PQG and PQ modules for feature decoding. The former generates easy and hard queries in stages to minimize false negatives, while the latter adaptively integrates dual-domain features to enhance regression accuracy of each query.
\end{enumerate}
\section{related work}
In this section, we overview homogeneous fusion methods in BEV and voxel domains. Additionally, since Mamba is utilized in this work, we also introduce recent advancements in state space models.

\subsection{Homogeneous Fusion in BEV Domain}
The introduction of homogeneous BEV fusion marks a significant milestone in the evolution of perception techniques for autonomous driving. A critical step in this method is the transformation of image features into the BEV space. The early study \cite{Yoo20203DCVFGJ} proposes an auto-calibration mechanism: image features are assigned to predefined voxel centers through 2D-3D projection and offset adjustment. These features are then compressed along the height dimension to generate BEV features. Building on this, \cite{tao_multimodal_2023} further introduces range attention to refine the distribution of intermediate voxel features along the range dimension. Recently, LSS \cite{Huang2021BEVDetHM} has become widely adopted for view transformation. It projects features into a 3D frustum based on depth estimation and then onto the BEV plane with a pooling operation. However, inaccuracies in depth estimation often lead to misalignment between LiDAR and image BEV features. To address this, \cite{Song2024ContrastAlignTR} leverages contrastive learning for feature alignment, while \cite{Fu2024EliminatingCC} and \cite{Wang2024TowardRL} utilize semantic-guided optical flow estimation or mutual deformable attention for explicit alignment. Additionally, \cite{Hu2023EALSSEL} and \cite{Wang2024TowardRL} attempt to enhance depth estimation accuracy through edge-aware LiDAR depth maps or local graph alignment in the frontal view.

Unlike the local fusion operations in the aforementioned methods, our approach establishes global relationships both within and across modalities, enabling effective feature alignment and a more comprehensive understanding of the driving scene.

\subsection{Homogeneous Fusion in Voxel Domain}
Compared to BEV homogeneous fusion, transforming image features into the voxel domain and fusing them with LiDAR voxels avoids the information loss caused by height compression. Additionally, introducing multi-modal voxel features during the decoding process leads to better regression accuracy. \cite{Li2025Multimodal3O} and \cite{Wu2022SparseFD} unproject image features from each pixel onto virtual points with depth completion and then convert them into voxels, applying voxel pooling to fuse fine-grained multi-modal features. \cite{Yin2021MultimodalVP} generates virtual points using nearest neighbor matching within 2D instance masks on the image. \cite{Jiao2022MSMDFusionFL} improves this approach with k-nearest neighbor matching and introduces gated modality-aware convolution to fuse semantic and geometric features of both camera and LiDAR at different granularities. While these virtual-point-based methods largely address the sparsity issue of LiDAR point clouds, they require extra depth completion or high-resolution instance segmentation networks. This leads to significant training costs and labels required for these tasks are often difficult to obtain. In contrast, \cite{Li2022UnifyingVR} bypasses the need for extra networks by adopting an LSS-like algorithm. It assigns image features to predefined dense voxels and then weights them with category-based depth maps. \cite{Ahmad2023mmFUSIONMF} eliminates the depth-weighting step and introduces additional gating mechanisms to enable flexible fusion. However, dense voxel representations incur significant computational overhead. In recent work, \cite{Song2024BiCoFusionBC} only utilizes element-wise fusion in the intermediate voxel space during LSS, with subsequent feature propagation occurring in the BEV domain.

In our work, we adopt sparse voxel representation to conserve computational resources. Additionally, our Mamba-based network—HVF, facilitates multi-granularity global feature fusion.

\subsection{State Space Models}
State Space Models (SSMs) \cite{Gu2021EfficientlyML, Gupta2022DiagonalSS,Gu2021CombiningRC} describe dynamic systems by representing the system's internal state at each moment. With advancements in deep learning, SSMs have been integrated into neural networks to achieve more efficient and flexible inference. \cite{Gu2020HiPPORM} effectively captures long-term dependencies in sequences by introducing higher-order polynomial projection operators, compressing historical information into a low-dimensional state space. S4 \cite{Gu2021EfficientlyML}, a structured state space model based on \cite{Gupta2022DiagonalSS}, enhances computational efficiency and modeling capacity through structured state transition matrices. Building on S4, \cite{Gu2023MambaLS} introduces an input-specific state transition mechanism--Mamba, significantly boosting both expressiveness and efficiency. Due to its linear complexity and modeling strength comparable to Transformer, many studies have explored its application in computer vision. For instance, \cite{Zhu2024VisionME} unfolds images into 1D sequences and proposes a bidirectional Mamba module for comprehensive feature learning. \cite{Liu2024VMambaVS} extends this approach to four-directional scanning, uncovering richer spatial relationships in images. It also reveals the connection between Mamba and linear attention through theoretical analysis, providing deeper insights into its underlying mechanisms. Mamba has also been applied to 3D tasks such as point cloud classification \cite{Liang2024PointMambaAS}, 3D object detection \cite{Zhang2024VoxelMG,Liu2024LIONLG,You2024MambaBEVAE} and semantic completion \cite{Tian2024MambaOccVS}. For example, \cite{Zhang2024VoxelMG} employs a group-free voxel Mamba module to expand the receptive field to the entire scene, while \cite{Tian2024MambaOccVS} integrates local adaptive reordering into the Mamba module to enhance local information extraction.

Different from these single-modality architectures for 3D tasks, we adapt Mamba to the multi-modal framework, further unlocking its potential of versatile applications.

\section{Method}
\subsection{Overview}
\begin{figure*}[tbp]
	\centering
	\includegraphics[width=\textwidth]{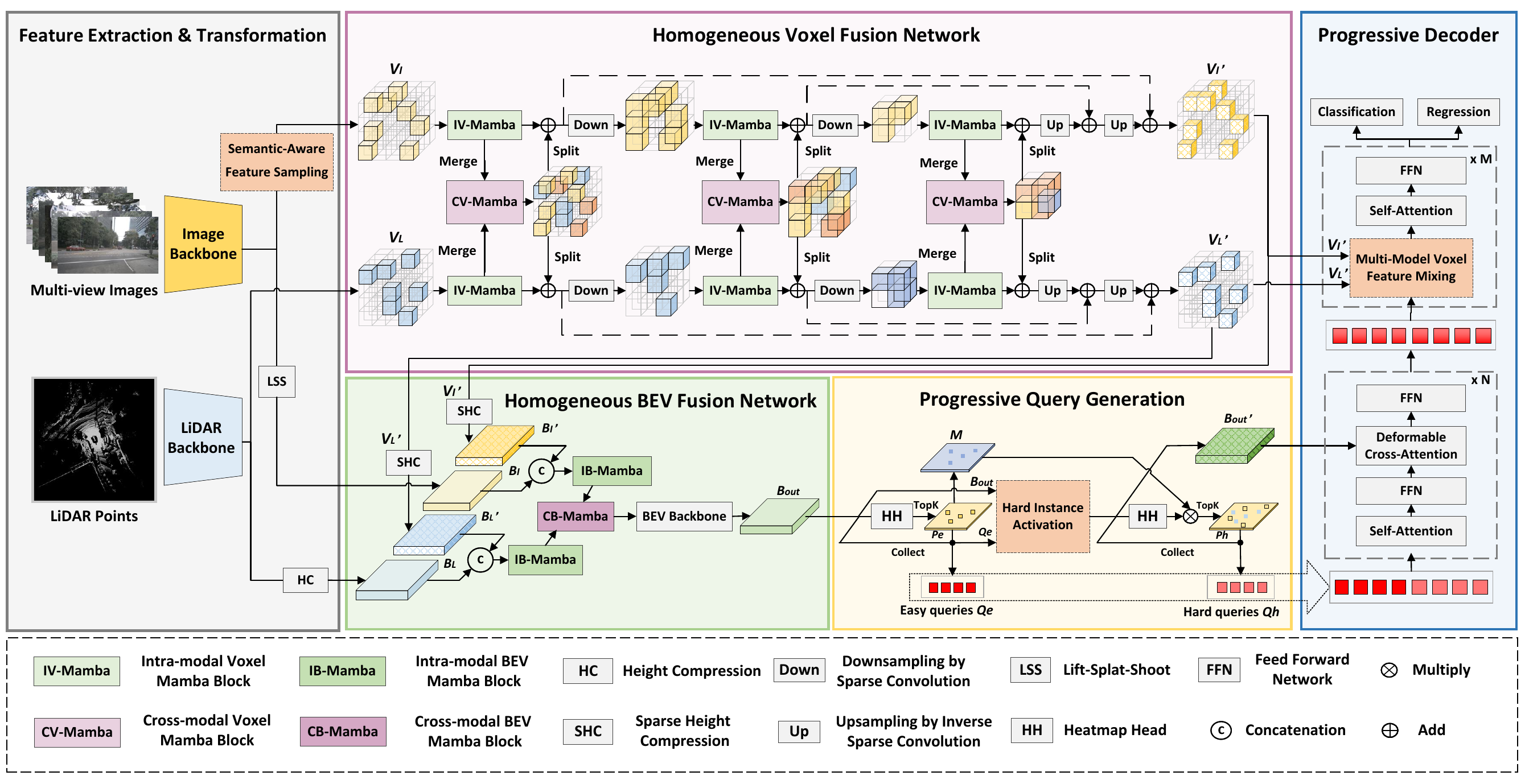} 
	\caption{Overview of DDHFusion. It begins by extracting features from multi-view images and LiDAR points and transforms both into the voxel and BEV domains. These features are then passed to two Mamba-based homogeneous fusion networks, which perform feature alignment and global perception. The resulting high-quality fused BEV feature $B_{out}$ is then fed into the progressive query generation module, which leverages the spatial relationship of easy queries to stimulate the generation of hard queries. Finally, all queries are passed into the progressive decoder, which abstracts dual-domain features around instances for accurate classification and box regression.} 
	\label{fig:pipeline}
\end{figure*}

The pipeline of DDHFusion is shown in Fig. \ref{fig:pipeline}. Initially, multi-view images and LiDAR points are passed through their respective backbones for feature extraction. The LiDAR voxels $V_L$ are converted  into $B_L$ through height compression, while image features are transformed into sparse voxels $V_I$, BEV map $B_I$ via LSS and SAFS modules. Then, we meticulously design the homogeneous voxel fusion network, where intra-modal voxel Mamba and cross-modal voxel Mamba (IV-Mamba and CV-Mamba) blocks are alternately employed through merge-and-split operations to interact $V_I$ and $V_L$. The resulting voxels $V_I^{'}$ and $V_L^{'}$ are fed into the homogeneous BEV fusion network and converted into $B_L'$, $B_L'$ by sparse height compression \cite{Chen2023VoxelNeXtFS}. The multi-modal BEV features $B_L$, $B_I$, $B_L'$ and $B_I'$ are fused using intra-modal and cross-modal BEV Mamba blocks (IB-Mamba and IB-Mamba) and the result is then processed through the BEV backbone for deep feature extraction. The resulting $B_{out}$ is used for progressive query generation. Finally, both easy queries $Q_e$ and hard queries $Q_h$ are concatenated and input into the progressive decoder which includes both BEV and voxel decoders for classification and box regression. In the remainder of this section, we provide a detailed introduction to DDHFusion from the following aspects: semantic-aware feature sampling, voxel homogeneous network, BEV homogeneous network, progressive query generation, and progressive decoder.

\subsection{Semantic-Aware Feature Sampling}
Describing the entire driving scene using dense voxels is highly redundant. First, a substantial portion of voxels corresponds to empty space, where no objects exist. Second, in autonomous driving scenarios, foreground objects typically occupy only a small fraction of the overall scene. This redundancy not only consumes considerable computational resources but also diverts the network's modeling capacity to large, irrelevant areas, thereby hindering feature learning.

\begin{figure}[tbp]
	\centering
	\includegraphics[width=0.5\textwidth]{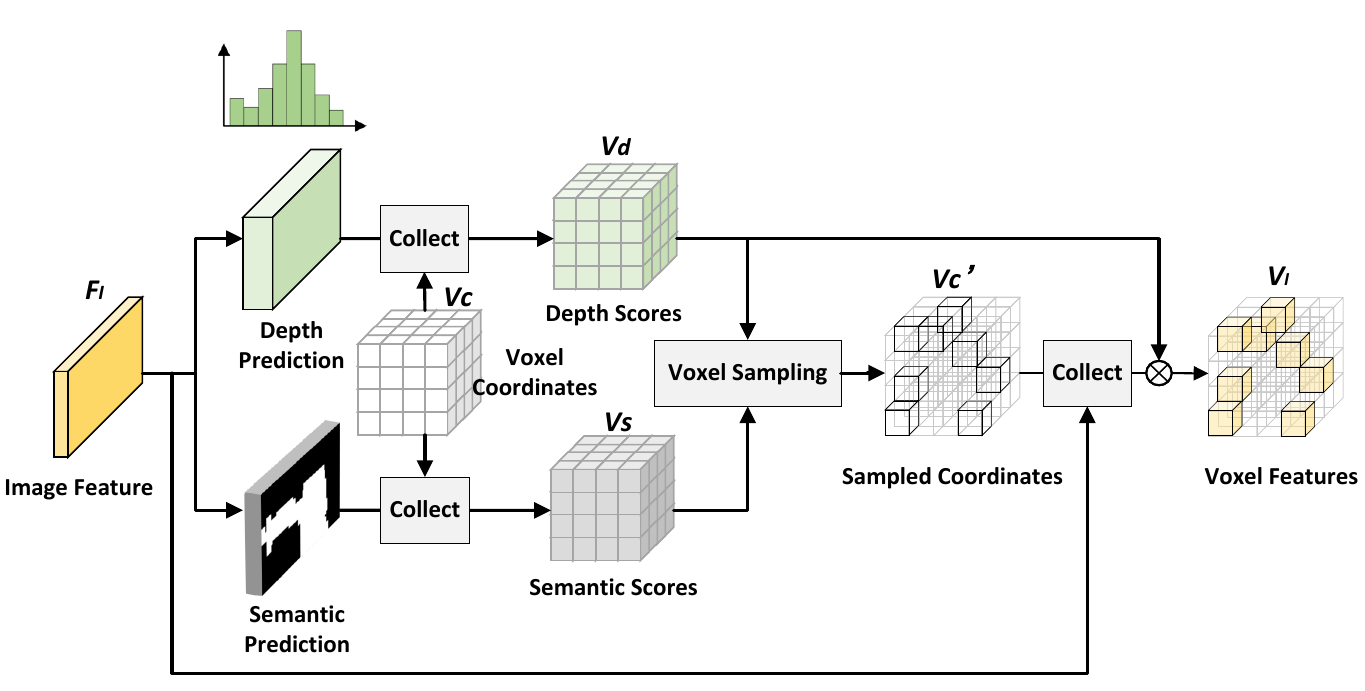} 
	\caption{Details of semantic-aware feature sampling (SAFS). We filter the coordinates of unimportant voxels by depth and semantic scores and transform the image feature to the sparse voxel domain.} 
	\label{SAFS}
\end{figure}

Drawing inspiration from \cite{Zhang2023SABEVGS}, we propose the SAFS module depicted in Fig. \ref{SAFS} to selectively generate important image voxels. First, the image feature $F_I$ from the image backbone are passed through a convolution block to generate the depth map. Similar to LSS, these depths are represented by discrete bins. Another branch predicts the semantic segmentation mask. During training, we project  points within ground truth boxes onto the image plane to generate segmentation labels for sparse supervision.

We divide the entire 3D space into voxels $V_c$ with number $H_I \times W_I \times Z_I$.  According to \cite{Li2022UnifyingVR}, we set the height range $Z_I$ twice as that of LiDAR voxels $Z_L$ to obtain more texture details from images. Then, we project voxel centers onto the image plane and collect depth scores $V_d$ and semantic score $V_s$ using bilinear interpolation. The voxels with both scores above certain thresholds are selected:

\begin{equation}
	V_c^{'} = \{ {v_{ci}} \mid ({v_{si}} > s) \land ({v_{di}} > d) | \notag 
\end{equation}
\begin{equation}
	\quad {v_{ci}} \in {V_c}, \, {v_{di}} \in {V_d}, \, {v_{si}} \in {V_s}, \, i \in [0, {H_I}{W_I}{Z_I} - 1] \}
\end{equation}
At the beginning of training or in some crowd scenarios, the number of $V_c^{'}$ may be too large. Therefore, we set an upper limit $N$. If the number exceeds it, the farthest point sampling is applied to retain only $N$ voxels. Finally, we assign image features to $V_c^{'}$ and multiply them by the corresponding depth scores to obtain voxel features $V_I$.

\subsection{Homogeneous Voxel Fusion Network}
\begin{figure*}[tbp]
	\centering
	\includegraphics[width=\textwidth]{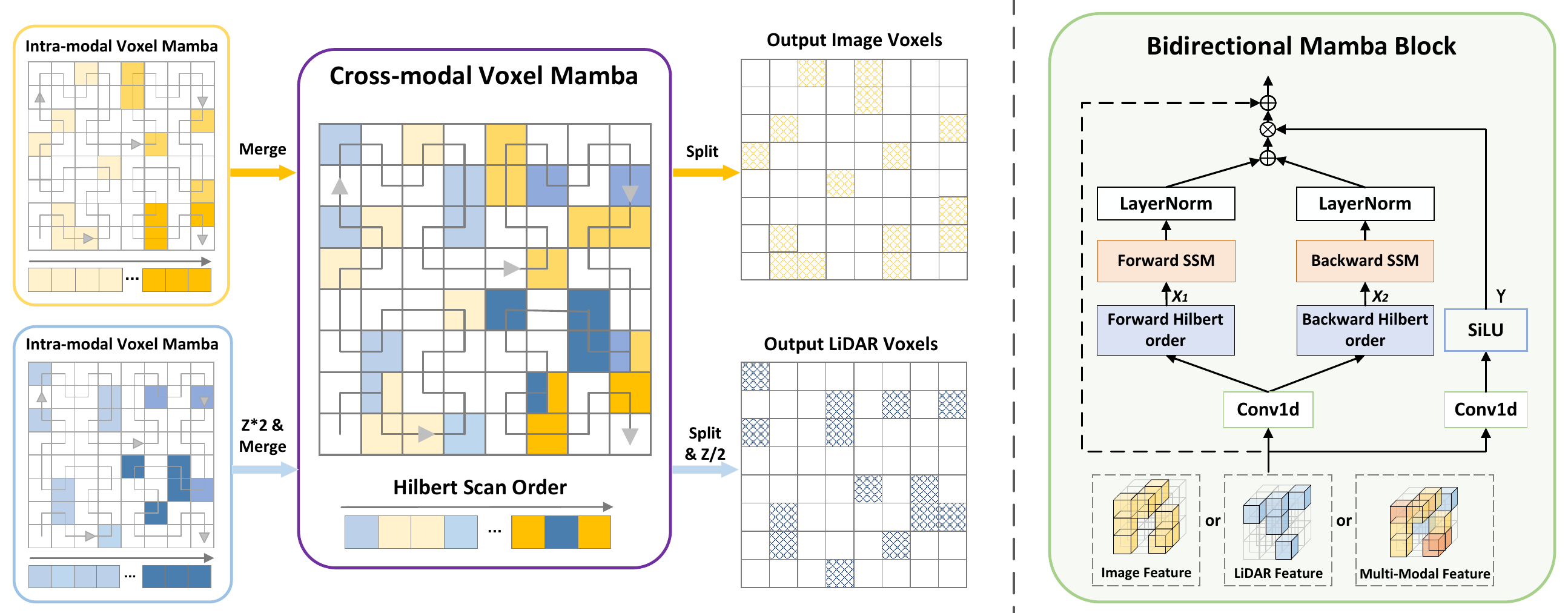} 
	\caption{Details of Mamba-based Voxel Fusion. In the left image, for the purpose of visualization, we use a 2D illustration to represent 3D scanning patterns in the HVF. "Z*2" and "Z/2" denote doubling or halving the z indices of LiDAR voxels (\textcolor{blue}{blue}) to align them with image voxels (\textcolor{orange}{yellow}) or to recover them to their original coordinates. As illustrated in the right figure, we utilize the bidirectional Mamba block in both intra-modal and cross-modal voxel Mamba modules.}
	\label{voxel_mamba}
\end{figure*}

As illustrated in Fig. \ref{voxel_mamba}, the HVF consists of two parallel branches that process voxel features separately, with feature fusion performed at each scale. Existing works \cite{Li2022UnifyingVR, Li2025Multimodal3O, Jiao2022MSMDFusionFL} fuse multi-modal voxel features through element-wise addition or local gating mechanisms limited to a small neighborhood. However, due to the uncertainty of depth estimation during feature transformation, the corresponding latent representations in $V_L$ and $V_I$ are often spatially deviated. With limited search ranges, these local fusion operations fail to effectively address this misalignment problem. Inspired by the recent success of Mamba in LiDAR-based 3D object detection \cite{Zhang2024VoxelMG,Liu2024LIONLG}, we propose IV-Mamba and CV-Mamba modules in the HVF for global feature interaction, which can better correlate severely misaligned multi-modal information and enable holistic understanding of the scene.

In IV-Mamba, following \cite{Zhang2024VoxelMG}, we rearrange the voxels into a 1D sequence in the order of the 3D Hilbert curve for its preservation of spatial proximity. The discrete SSM model is then used to process the entire sequence:

\begin{equation}
	\left\{
	\begin{array}{l}
		h_i = \overline{A} h_{i - 1} + \overline{B} x_i \\
		y_{i - 1} = \overline{C} h_i + x_i
	\end{array}
	\right.
	\label{ssm}
\end{equation}
\begin{figure*}[htbp]
	\centering
	\includegraphics[width=\textwidth]{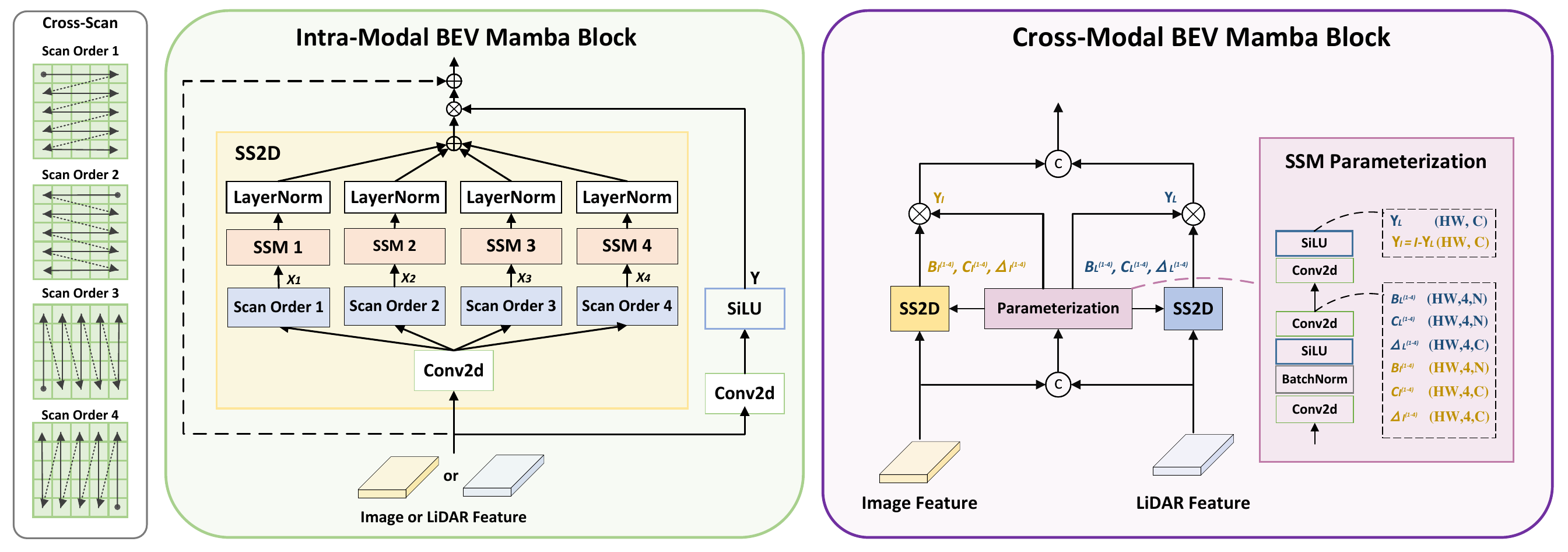} 
	\caption{Details of Intra-Modal and Cross-Modal BEV Mamba. We apply the four-directional scanning to unfold the BEV feature into 1D sequences, which helps construct a comprehensive spatial relationship. In the cross-modal Mamba block, the parameters for the SS2D modules of each modality are jointly computed from the image and LiDAR BEV features, while $Z_I$ and $Z_L$ are used to adaptively modulate the weights of their respective modalities.} 
	\label{bev_mamba}
\end{figure*}
Here, $\overline{A}$ and $\overline{B}$ are derived from the parameters of the continuous SSM using the zero-order hold formula:

\begin{equation}
	\left\{ {\begin{array}{*{20}{c}}
			{\overline {A}  = \exp (\Delta {A})}\\
			{\overline {B} = {{(\Delta {A)}}^{ - 1}}(\exp (\Delta {A}) - {I}) \cdot {B}}
	\end{array}} \right.
	\label{ssm2}
\end{equation}
Where $\bigtriangleup$ denotes the time-scale parameter and $A$ denotes the learnable matrix. $B$, $C$, and $\bigtriangleup$ are inferred from the input sequence $x_i$ through linear layers, ensuring the input-specific of the SSM. To ensure complete perception, both forward and backward SSMs are implemented. Their output sequences are added together and modulated by $Y$ to produce the final results.

In CV-Mamba, we directly combine multi-modal voxels in the unified 3D coordinates. It’s important to note that the height range of image voxels $Z_I$ is twice that of LiDAR voxels $Z_L$. Therefore, we need to double the z indices of LiDAR voxels before combination. Similar to IV-mamba, these voxels are unfolded into a 1D sequence and fed into the SSM. While some voxels may occupy the same position, we treat them as distinct elements in the sequence rather than merging them as in \cite{Jiao2022MSMDFusionFL}. Then, the output voxels with different modalities are separated and sent back to their respective branches, where they are downsampled by sparse 3D convolution. Notably, in most existing methods, fusing sparse features with different distributions requires operations such as neighbor search \cite{Wu2022SparseFD,Jiao2022MSMDFusionFL,Li2025Multimodal3O} or equal-sized grouping  \cite{Wang2023UniTRAU}, while our designed elegant global merge-and-split operation makes feature fusion more straightforward and efficient. Alternately processing voxels with IV-Mamba and CV-Mamba effectively bridges the modality gap and alleviates the negative impact of deteriorated data from each modality.

Finally, inverse sparse convolutions are used to upsample multi-scale voxel features back to the original scale. The output features $V_I^{'}$ and $V_L^{'}$ are not directly combined to avoid interference caused by distinctive distributions. They are fed into the HBF network and the MMVFM module in the decoder.

\subsection{Homogeneous BEV Fusion Network}
In the HBF Network, sparse height compression \cite{song_voxelnextfusion_2023} is utilized to transform  $V_I^{’}$ and $V_L^{’}$ into BEV features. Specifically, It applies max-pooling to voxel features belonging to the same BEV pixel, yielding $B_I^{'}$ and $B_L^{'}$. They are then concatenated with $B_I$ and $B_L$ to inject more 3D texture and structure details into the BEV space. It is important to note that during SAFS, some potential foreground voxels may be filtered out due to low semantic or depth scores. The LSS compensates for this by retaining image features of the entire environment. Moreover, LSS and SAFS share the same depth map in our work. Since LSS doesn't undergo feature filtering, it preserves all backpropagated gradients during depth estimation, which indirectly accelerates the convergence of SAFS during training.

Like HVF, we also apply Mamba-based feature fusion including both IB-Mamba and CB-Mamba blocks to generate high-quality BEV features. Firstly, the concatenated features are input into IB-Mamba to globally perceive modality-specific information. Following the SS2D block proposed in \cite{Liu2024VMambaVS}, we apply the four-directional cross-scanning to flatten BEV tensors, which helps construct a comprehensive spatial relationship. As described in Equations \ref{ssm} and \ref{ssm2}, the SSM is performed simultaneously across all four directions. After passing through LayerNorm, the outputs are summed and modulated by $Y$ for the final result.

In CB-Mamba, to further align multi-modal features in the dense BEV space and adaptively fuse them, we extend the IB-mamba to accommodate dual inputs for cross-modal fusion. As shown in Fig. \ref{bev_mamba}, the parameters for both LiDAR and image SS2D are generated from the concatenated multi-modal BEV features, guiding the feature propagation for both image and LiDAR features. According to the theoretical proof in \cite{Liu2024VMambaVS}, the SS2D is a special variant of the global transformer. Therefore, CB-Mamba can match corresponding elements between modalities for feature alignment. We describe the SSM parameterization with the  following formulas:
\begin{equation}
	T = Conv2d(SiLU(BatchNorm(Conv2d(F_{comb}))))
\end{equation}
\begin{equation}
	[B_I^{1-4},C_I^{1-4},\bigtriangleup_I^{1-4},B_L^{1-4},C_L^{1-4},\bigtriangleup_L^{1-4}] = Split(T)
\end{equation}
Where, the superscript $1-4$ denotes the parameters employed for the four-directional scanning in SS2D. Furthermore, the modulation parameters $Y_I$ and $Y_L$ are also generated to reweight multi-modal features:

\begin{equation}
	Y_I = SiLU(Conv2d(T))
\end{equation}
\begin{equation}
	Y_L = I - Y_I
\end{equation}
It enables the network to autonomously favor the more reliable modality, thereby improving adaptability in complex scenarios‌. 

After CB-Mamba, the output BEV tensor is fed into the BEV backbone, yielding $B_{out}$, which is then utilized as the input of the PQG module.

\begin{figure}[t]
	\centering
	\includegraphics[width=0.5\textwidth]{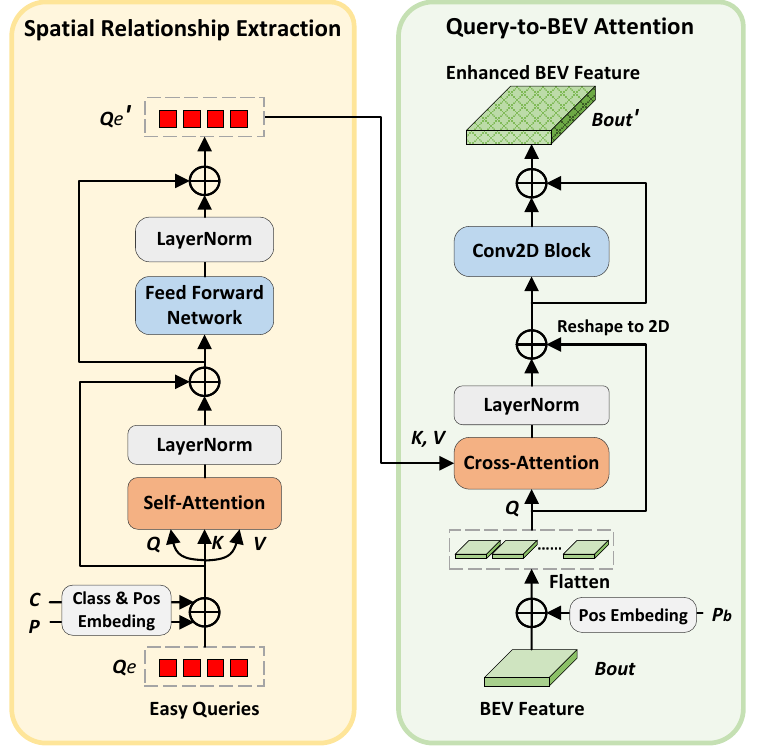} 
	\caption{Details of Hard Instance Activation. First, self-attention among easy queries is used to extract spatial relationships between objects. Then, query-to-BEV attention is applied to activate latent hard instances within the BEV feature.} 
	\label{HIA}
\end{figure}

\subsection{Progressive Query Generation}
In DDHFusion, the BEV domain is used for efficient generation of target candidates. While many previous works have attempted to achieve this in the range view\cite{fan_rangedetdefense_2021,meng_small_2024} or sparse 3D space\cite{Yang20203DSSDP3, fan_fully_2022}, the former is impacted by object occlusions and the latter often relies on center-voting strategies, which are sensitive to uneven voxel or point densities. In contrast, selecting object queries in the BEV domain is relatively simple. First, we generate a heatmap from $B_{out}^{’}$ using a convolutional operation. As described in \cite{Bai2022TransFusionRL}, after processing the heatmap using non-maximum suppression based on 3x3 max pooling, we can obtain queries via the top-k selection. However, there remain some hard targets that tend to be ignored during selection. They often exhibit corrupt information such as sparse surface points or blurred image textures. After height compression, their features may degrade further. Moreover, since targets are typically small in the BEV plane, their features may be easily overshadowed during convolution operations in the BEV backbone.

To address this, \cite{Chen2023FocalFormer3DF} propose a multi-stage query generation strategy. It groups queries into different difficulty levels and employs residual blocks to activate BEV features between levels to focus on harder instances. In this work, we simplify the approach by dividing queries into two groups: easy queries $Q_e$ and hard queries $Q_h$. The corresponding two-stage extraction process is detailed in the following equations:

\begin{equation}
	{H} = Heatmaphead({B_{out}})
\end{equation}
\begin{equation}
	{P_e} = Topk(NMS(H))
\end{equation}
\begin{equation}
	{Q_e} = Collect(B_{out},P_e)
\end{equation}
\begin{equation}
	M =	I - Maxpooling(Mask(P_e))
\end{equation}
\begin{equation}
	B_{out}^{'} = Resnet({B_{out}})
	\label{res1}
\end{equation}
\begin{equation}
	H^{'} = Heatmaphead({B_{out}^{'}})
	\label{res2}
\end{equation}
\begin{equation}
	P_h = Topk(NMS(M\cdot{H^{'}}))
\end{equation}
\begin{equation}
	{Q_h} = Collect(B_{out}^{'},P_h)
\end{equation}
Where $P_e$ and $P_h$ denote the positions of easy and hard queries. $M$ represents the mask generated from $P_e$, which is used to obtain $P_h$ by masking the expanded regions corresponding to $Q_e$. During training, we need to compute the heatmap loss of the second stage that doesn't lie within the masked regions of $Q_e$, guiding the network to focus on hard targets. However, activating BEV feature by the residual block has limited effectiveness—it can only aggregate local information from difficult regions to enhance responses. In the driving environment, the spatial relationships between objects exhibit certain regularities, such as the placement of vehicles and barriers. Based on this observation, we propose a hard instance activation (HIA) module, which leverages easy queries $Q_e$ to activate features in difficult regions. Specifically, we modify Equation \ref{res1} as follows:

\begin{equation}
	{B_{out}^{'}} = HIA(\textcolor{red}{Q_e},{B_{out}})
\end{equation}
As shown in Fig \ref{HIA}, in HIA, we first embed categories -- in form of one-hot vectors -- and positions into the easy queries. Then, we utilize self-attention to extract relationships between objects, resulting in enhanced queries $Q_e^{'}$. Then, $B_{out}$ is flattened and regarded as the query and $Q_e^{'}$ serves as the key and value. The query-to-BEV attention\cite{Chen2024LearningHV},\cite{Yin2024ISFusionIC} is utilized to propagate object relationships to every pixel in the BEV map. Subsequently, the results are reshaped back into a 2D tensor and passed through a residual block to obtain the refined BEV features $B_{out}^{'}$ for the selection of hard queries $Q_h$. Finally, the easy and hard queries along with $B_{out}^{'}$ are fed into the progressive decoder .

\begin{figure}[t]
	\centering
	\includegraphics[width=0.5\textwidth]{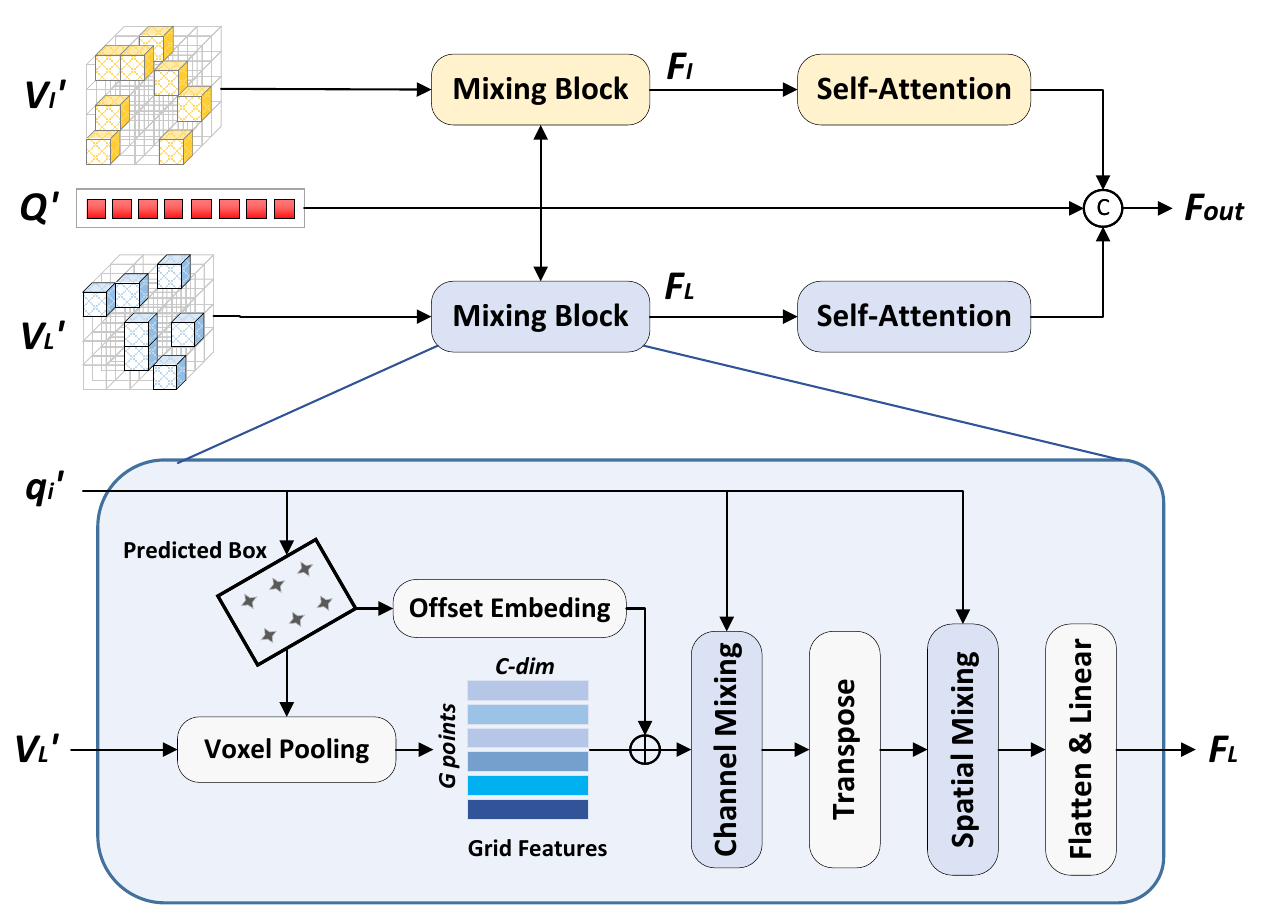} 
	\caption{In the mixing block, queries are used to dynamically aggregate nearby voxel features from each modality. Subsequently, self-attention is employed to extract long-range features among instances.} 
	\label{mix}
\end{figure}
\subsection{Progressive Decoder}
Many works \cite{liu_bevfusion_2023, Yin2024ISFusionIC, Chen2023FocalFormer3DF, Song2024GraphBEVTR} only decode the BEV features at the end of their models. Although these features are rich in contextual and high-level semantic information, they lack fine-grained geometry-aware details, which limits the accuracy of box regression. To solve this, our progressive decoder (PQ) incorporates features from both BEV and voxel domains.

As shown in Fig \ref{fig:pipeline}, we decode the BEV features with several deformable attention blocks to obtain the refined queries $q^{'}$. They are then fed into the (MMVFM) module to decode voxel features $V_I^{'}$ and $V_L^{'}$. As depicted in Fig \ref{mix}, two separate mixing modules are designed to dynamically aggregate voxel features. In each, we use several MLPs to regress the proposal box $b_i$ from the query $q_i^{'}$. $b_i$ is evenly divided into $G = g\times g \times g$ grid points $g_1,…g_G$. Next, we extract grid features $Fg=[f_{1},…,f_{G}]$ from $V_L$ or $V_I$ using voxel pooling. We then embed offsets $\bigtriangleup g_{xk},\bigtriangleup g_{yk},\bigtriangleup g_{zk}$ of each grid point relative to the center of $b_i$ into $F_g$ through a linear layer, generating  shape-aware grid features $F_g^{'}\in \mathbb{R}^{G\times C}$.  Following Adamixer \cite{Gao2022AdaMixerAF}, we sequentially perform channel mixing and spatial mixing on them:
\begin{equation}
	C_i = Linear(q_i)
\end{equation}
\begin{equation}
	F_g^{'} = F_g \times {C_i}
\end{equation}
\begin{equation}
	S_i = Linear(q_i)
\end{equation}
\begin{equation}
	F_g^{''} = transpose(F_g^{'}) \times {S_i}
\end{equation}
Here, \( C_i \in \mathbb{R}^{C \times C} \) and \( S_i \in \mathbb{R}^{G \times G/4} \) represent the channel and spatial kernels, respectively. After the mixing operation, \( F_g'' \) is flattened into a vector and passed through a linear layer to reduce the dimensionality, resulting in the output \( F_I \) or \( F_L \). It is worth noting that \( S_i \) differs in input and output channels from the original design of Adamixer \(\mathbb{R}^{G \times G} \). Ours can reduce the number of parameters by compressing the spatial dimension. Following the aforementioned local mixing operation, we further aggregate long-range 3D features between objects with self-attention. Finally, the features from both modalities, along with queries, are concatenated as the output. Subsequently, we decode the output features through self-attention and a feed-forward layer. Finally, a detection head is applied to regress confidences and bounding boxes.

\section{EXPERIMENT}
Our method is evaluated on the challenging NuScenes dataset. In this section, we first describe the dataset and implementation details, then report the evaluation performance of DDHFusion. Furthermore, we conduct ablation studies of each component and comparison experiments with other homogeneous fusion methods for a thorough analysis.
\subsection{Experimental Setup}
\noindent\textbf{Dataset.} The nuScenes dataset is notable for its multi-sensor setup, which includes LiDAR, camera, radar, and GPS/IMU data, enabling comprehensive multi-modal research. It comprises 1,000 diverse driving scenarios, with a standardized split of 700 scenes for training, 150 for validation, and 150 for testing. It also features 10 object categories, such as cars, pedestrians, and cyclists, with over 1.4 million 3D bounding boxes annotated across various driving conditions. For 3D object detection, two key metrics are employed: mean Average Precision (mAP) and the nuScenes Detection Score (NDS), which collectively provide a holistic assessment of model performance. The mAP is computed by averaging the precision across 10 categories at distance thresholds of 0.5m, 1m, 2m, and 4m. NDS, on the other hand, is a composite metric that combines mAP with five True Positive (TP) error measures: mean Average Translation Error (mATE), mean Average Scale Error (mASE), mean Average Orientation Error (mAOE), mean Average Velocity Error (mAVE), and mean Average Attribute Error (mAAE). 
\begin{table*}[tbp]
	\centering
	\renewcommand{\arraystretch}{1.2} 
	\caption{Evaluation results on the nuScenes test set. All models in the table are without ensemble or test-time augmentation. “L” denotes LiDAR and “C” denotes camera.}
	\label{tab:test}
	\begin{tabular}{l|c|c|c|c|c|c|c|c|c|c|c|c|c}
		\hline
		Method & Modality & mAP & NDS & Car & Truck & C.V. & Trailer & Bus & Barrier & Motor. & Bike & Ped. & T.C. \\
		\hline
		\hline
		CenterPoint \cite{Yin2020Centerbased3O} & L & 60.3 & 67.3 & 85.2 & 53.5 & 20.0 & 56.0 & 63.6 & 71.1 & 59.5 & 30.7 & 84.6 & 78.4 \\
		TransFusion-L \cite{Bai2022TransFusionRL} & L & 65.5 & 70.2 & 86.2 & 56.7 & 28.2 & 58.8 & 66.3 & 78.2 & 68.3 & 44.2 & 86.1 & 82.0 \\
		LargeKernel3D \cite{Chen2022LargeKernel3DSU} & L & 65.3 & 70.5 & 85.9 & 55.3 & 26.8 & 60.2 & 66.2 & 74.3 & 72.5 & 46.6 & 85.6 & 80.0 \\
		FocalFormer3D \cite{Chen2023FocalFormer3DF} & L & 68.7 & 72.6 & 87.8 & 59.4 & 37.8 & 65.7 & 73.0 & 77.8 & 77.4 & 52.4 & 90.0 & 83.4 \\
		\hline
		MVP \cite{Yin2021MultimodalVP} & LC & 66.4 & 70.5 & 86.8 & 58.5 & 26.1 & 57.3 & 67.4 & 74.8 & 70.0 & 49.3 & 89.1 & 85.0 \\
		AutoAlignV2 \cite{Chen2022AutoAlignV2DF} & LC & 68.4 & 72.4 & 87.0 & 59.0 & 33.1 & 59.3 & 69.3 & - & 72.9 & 52.1 & 87.6 & - \\
		MSMDFusion \cite{Jiao2022MSMDFusionFL} & LC & 71.5 & 74.0 & 88.4 & 61.0 & 35.2 & 66.2 & 71.4 & 80.7 & 76.9 & 58.3 & 90.6 & 88.1 \\
		TransFusion \cite{Bai2022TransFusionRL} & LC & 68.9 & 71.7 & 87.1 & 60.0 & 33.1 & 60.8 & 68.3 & 78.1 & 73.6 & 52.9 & 88.4 & 86.7 \\
		BEVFusion \cite{liu_bevfusion_2023} & LC & 70.2 & 72.9 & 88.6 & 60.1 & 39.3 & 63.8 & 69.8 & 80.0 & 74.1 & 51.0 & 89.2 & 85.2 \\
		BEVFusion \cite{liang_bevfusion_2022} & LC & 71.3 & 73.3 & 88.1 & 60.9 & 34.4 & 62.1 & 69.3 & 78.2 & 72.2 & 52.2 & 89.2 & 86.7 \\    
		UVTR \cite{Li2022UnifyingVR} & LC & 67.1 & 71.1 & - & - & - & - & - & - & - & - & - & - \\
		GraphBEV \cite{Song2024GraphBEVTR} & LC & 71.7 & 73.6 & 89.2 & 60.0 & \textbf{40.8} & 64.5 & 72.1 & 80.1 & 76.8 & 53.3 & \textbf{90.9} & \textbf{88.9} \\    
		FocalFormer3D \cite{Chen2023FocalFormer3DF} & LC & 71.6 & 73.9 & 88.5 & 61.4 & 35.9 & 66.4 & 71.7 & 79.3 & 80.3 & 57.1 & 89.7 & 85.3 \\
		CMT \cite{Yan2023CrossMT} & LC & 70.4 & 73.0 & 88.0 & 63.3 & 37.3 & 65.4 & \textbf{75.4} & 78.2& 79.1& 60.6 &87.9& 84.7 \\    
		ObjectFusion \cite{Cai2023ObjectFusionM3} & LC & 71.0 & 73.3 & 89.4 & 59.0 & 40.5 & 63.1 & 71.8 & 76.6 & 78.1 & 53.2 & 90.7 & 87.7 \\
		DeepInteraction \cite{Chen2022AutoAlignV2DF} & LC & 70.8 & 73.4 & 87.9 & 60.2 & 37.5 & 63.8 & 70.8 & 80.4 & 75.4 & 54.5 & 91.7 & 87.2 \\
		DeepInteraction++ \cite{Yang2024DeepInteractionMI} & LC & 72.0 & 74.4 & - & - & - & - & - & - & - & - & - & - \\
		BiCo-Fusion \cite{Song2024BiCoFusionBC} & LC & 72.4 & 74.5 & 88.1 & 61.9 & 38.2 & 65.7 & 73.3 & 80.4 & 78.9 & 59.8 & 89.7 & 88.3 \\
		DDHFusion(ours) & LC & \textbf{73.8} & \textbf{75.6} & \textbf{90.4} & \textbf{63.6} & 39.1 & \textbf{66.7} & 74.4 & \textbf{81.0} & \textbf{81.8} & \textbf{63.6} & 90.3 & 87.5 \\
		\hline
	\end{tabular}
\end{table*}

\noindent\textbf{Implementation Details.} Our model is implemented on the MMDetection3D framework.  For the LiDAR backbone, we employ VoxelNet~\cite{Yan2018SECONDSE}, while Swin-T serves as the image backbone. The detection range is set to $[-54m, 54m]$ along x and y axes, and $[-5m, 3m]$ in the z direction. For ablation studies and extensive comparative experiments, the image size is set to $256\times704$ for fast training. When evaluating the model on the validation or test sets, we follow previous works to set the image size to $384\times1056$ and double the kernel channels in the LiDAR backbone to achieve better performance. For SAFS, we set score thresholds $d=0.01$, $s=0.25$. The maximum voxel number $N$ is set to $18000$. DDHFusion is trained on four NVIDIA 4090D GPUs with batch size of $12$. The training process is divided into two stages: (1) training the LiDAR-only network for 20 epochs. (2) combining the image and LiDAR branches and training the fused network for 6 epochs.

During training, we follow CBGS [51] to perform class-balanced sampling and employ the AdamW
optimizer with maximum learning rate $2e^{-3}$. The copy-paste data augmentation \cite{Yan2018SECONDSE} is applied to prevent overfitting. 

\begin{table}[htbp]
	\centering
	\caption{Evaluation results on the nuScenes val set. “L” denotes LiDAR and “C” denotes camera.}
	\label{tab:val}
	\renewcommand{\arraystretch}{1.2} 
	\begin{tabular}{l|c|c|c}
		\hline
		Methods & Modality & mAP$\uparrow$ & NDS$\uparrow$ \\
		\hline
		\hline
		CenterPoint \cite{Yin2020Centerbased3O} & L & 60.3 & 67.3 \\
		TransFusion-L \cite{Bai2022TransFusionRL} & L & 65.5 & 70.2 \\
		TransFusion \cite{Bai2022TransFusionRL} & LC & 67.5 & 71.3 \\
		BEVFusion \cite{liu_bevfusion_2023} & LC & 68.5 & 71.4 \\
		GraphBEV \cite{Song2024GraphBEVTR} & LC & 70.1 & 72.9 \\
		CMT \cite{Yan2023CrossMT} & LC & 67.9 & 70.8 \\
		MSMDFusion \cite{Jiao2022MSMDFusionFL} & LC & 69.3 & 72.1 \\
		ObjectFusion \cite{Cai2023ObjectFusionM3} & LC & 69.8 & 72.3 \\
		UVTR \cite{Li2022UnifyingVR} & LC & 65.4 & 70.2 \\
		DeepInteraction \cite{Chen2022AutoAlignV2DF} & LC & 69.9 & 72.6 \\
		DeepInteraction++ \cite{Yang2024DeepInteractionMI} & LC & 70.6 & 73.3 \\
		BiCo-Fusion\cite{Song2024BiCoFusionBC} & LC & 70.5 & 72.9 \\
		DDHFusion (ours) & LC & 71.6 & 73.8 \\
		DDHFusion-large (ours) & LC & \textbf{72.8} & \textbf{74.4} \\
		\hline
	\end{tabular}
\end{table}

\subsection{Results and Comparison}
As shown in Table \ref{tab:test}, we report the results of DDHFusion on the NuScenes test set. It achieves an impressive 73.8 mAP and 75.6 NDS without model ensemble or test-time augmentation, surpassing all other state-of-the-art methods. Compared to the second-best Bico-Fusion model, DDHFusion shows improvements of 1.4 mAP and 1.1 NDS. In Table \ref{tab:val}, we also present the performance of DDHFusion on the NuScenes validation set. DDHFusion outperforms the second-best DeepInteraction++ model by 1.0 mAP and 0.5 NDS. By increasing parameters and expanding the input image size, DDHFusion-Large achieves even greater accuracy advantages. These improvements are attributed to the complementary strengths voxel and BEV homogeneous domains, as well as advanced mechanism that we propose to provide larger receptive fields, fewer false negatives and higher localization accuracy.

\begin{table}[H]
	\centering
	\renewcommand{\arraystretch}{1.2} 
	\caption{Ablation study of each module.}
	\label{tab:Ablation Studies}
	\begin{tabular}{c|c|c|c|c|c|c}
		\hline
		Methods & HVF & HBF & PQG & PD & mAP & NDS \\
		\hline
		\hline
		(1) & & & & & 68.4 & 71.2 \\
		(2) & \checkmark & & & & 70.0 & 72.3 \\
		(3) & & \checkmark & & & 69.1 & 71.8 \\
		(4) & \checkmark & \checkmark & & & 70.8 & 72.7 \\
		(5) & \checkmark & \checkmark & \checkmark & & 71.4 & 73.1 \\
		(6) & \checkmark & \checkmark & \checkmark & \checkmark & \textbf{71.6} & \textbf{73.8} \\
		\hline
	\end{tabular}
\end{table}
\begin{table}[H]
	\centering
	\renewcommand{\arraystretch}{1.2} 
	\caption{Ablation study of HVF.}
	\label{tab:Voxel_mamba}
	\begin{tabular}{c|c|c|c|c}
		\hline
		Methods & IV-Mamba & CV-Mamba & mAP & NDS \\
		\hline
		\hline
		(1) & & & 68.4 & 71.2 \\
		(2) & \checkmark & & 69.4 & 71.9 \\
		(3) & & \checkmark & 69.6 & 71.8 \\
		(4) & \checkmark & \checkmark & \textbf{70.0} & \textbf{72.3} \\
		\hline
	\end{tabular}
\end{table}
\begin{table}[H]
	\centering
	\renewcommand{\arraystretch}{1.2} 
	\caption{Ablation study of different downsampling strides in HVF.}
	\label{tab:voxel_stride}
	\begin{tabular}{c|c|c|c}
		\hline
		Method & Downsampling strides & mAP & NDS \\
		\hline
		\hline
		(1) & [1] & 69.5 & 71.9 \\
		(2) & [1, 2] & 69.8 & 72.2 \\
		(3) & [1, 2, 4] & \textbf{70.0} & \textbf{72.3} \\
		\hline
	\end{tabular}
\end{table}
\begin{table}[H]
	\centering
	\renewcommand{\arraystretch}{1.2} 
	\caption{Ablation study of HBF.}
	\label{tab:bev_mamba}
	\begin{tabular}{c|c|c|c|c}
		\hline
		Methods & IB-Mamba & CB-Mamba & mAP & NDS \\
		\hline
		\hline
		(1) & & & 70.0 & 72.3 \\
		(2) & \checkmark & & 70.3 & 72.4 \\
		(3) & & \checkmark & 70.5 & 72.5 \\
		(4) & \checkmark & \checkmark & \textbf{70.8} & \textbf{72.7} \\
		\hline
	\end{tabular}
\end{table}

\subsection{Ablation Studies}
In this section, we demonstrate the effectiveness of each component through comprehensive ablation studies.

\noindent\textbf{Ablation of each module.} As shown in Table \ref{tab:voxel_stride}, we evaluate the contribution of each module to the detection performance. We replace the decoder in BEVFusion with deformable attention as the baseline. First, the proposed HVF significantly improves performance by 1.6 mAP and 1.1 NDS. Introducing only the HBF network also provides a gain of 0.7 mAP and 0.6 NDS. As reported in the third row, the combination of both homogeneous fusion networks achieves a substantial improvement of 2.4 mAP and 1.6 NDS. Additionally, the PQG module enhances detection recall, contributing 0.6 mAP and 0.4 NDS. Finally, integrating voxel decoder into the original BEV decoder to form the progressive decoder further increases performance by 0.2 mAP and 0.7 NDS. In summary, the four proposed components can fully exploit the potential of the dual-domain fusion framework for high-performance object detection.

\begin{table}[H]
	\centering
	\renewcommand{\arraystretch}{1.2} 
	\caption{Comparsion of different activation methods in PQG.}
	\label{tab:query_generation}
	\begin{tabular}{c|c|c}
		\hline
		Activation Methods & mAP & NDS \\
		\hline
		\hline
		Baseline & 70.8 & 72.7 \\
		Residual block\cite{Chen2023FocalFormer3DF} & 71.0 & 72.8 \\
		HIA (ours) & \textbf{71.4} & \textbf{73.1} \\
		\hline
	\end{tabular}
\end{table}
\begin{table}[H]
	\centering
	\renewcommand{\arraystretch}{1.2} 
	\caption{Comparsion of different decoder numbers in MMVFM.}
	\label{tab:voxel_mix}
	\begin{tabular}{c|c|c|c|c}
		\hline
		Methods & BEV decoder N & voxel decoder M & mAP & NDS \\
		\hline
		\hline
		(1) & 0 & 1 & 70.9 & 72.6 \\
		(2) & 3 & 0 & 71.4 & 73.1 \\
		(3) & 3 & 1 & \textbf{71.6} & 73.8 \\
		(4) & 3 & 2 & 71.5 & \textbf{73.9} \\
		(5) & 3 & 3 & 71.6 & 73.7 \\
		\hline
	\end{tabular}
\end{table}
\begin{table}[H]
	\centering
	\renewcommand{\arraystretch}{1.2} 
	\caption{Ablation study of MMVFM}
	\label{query_decoder}
	\begin{tabular}{c|c|c|c|c}
		\hline
		Methods & mixing block & self-attention& mAP & NDS \\
		\hline
		\hline
		(1) & & & 71.4 & 73.1 \\
		(2) & \checkmark & & 71.5 & 73.3 \\
		(3) & & \checkmark & 71.6 & 73.4 \\
		(4) & \checkmark & \checkmark & \textbf{71.6} & \textbf{73.8} \\
		\hline
	\end{tabular}
\end{table}
\begin{table}[H]
	\centering
	\renewcommand{\arraystretch}{1.2} 
	\caption{Analysis of scaling choices.}
	\label{scaling}
	\begin{tabular}{c|c|c}
		\hline
		Scaling choice & mAP & NDS \\
		\hline
		\hline
		Baseline & 71.6 & 73.8 \\
		\hline
		+ Doubling channels in LiDAR backbone  & 71.9 & 74.0 \\
		\hline
		+ Expanding input image size & \textbf{72.8} & \textbf{74.4} \\
		\hline
	\end{tabular}
\end{table}

\noindent\textbf{Analysis of homogeneous voxel fusion network.} As shown in Table \ref{tab:Voxel_mamba}, we analyze the role of the Mamba-based modules in the voxel fusion stage. In the second row, we introduce a 3D U-Net with only IV-Mamba, without directly fusing image and LiDAR voxels. It achieves a performance improvement of 1.0 mAP and 0.7 NDS, which can be attributed to IV-Mamba’s capability to effectively capture long-range dependencies within each modality. The CV-Mamba can also enhance network performance, indicating the effectiveness of global fusion of voxel features with different modalities and distributions. By combining both IV-Mamba and CV-Mamba with the merge-and-split operation, we not only align multi-modal features but also preserve the relative independence of each modality, avoiding interference from noise or low-quality information from other modality, thereby further improving accuracy. As shown in Table~\ref{tab:voxel_stride}, setting the downsampling strides to $[1, 2, 4]$ in HVF can achieve the best performance, as increasing the downsampling rate allows the network to learn structural information at different granularities.

\noindent\textbf{Analysis of homogeneous BEV fusion network} We further investigate HBF. As shown in Table \ref{tab:bev_mamba}, the SS2D module in IB-Mamba enables global feature diffusion across the dense BEV map in four directions, resulting in improvements of 0.3 mAP and 0.1 NDS. Our proposed CB-Mamba achieves gains of 0.5 mAP and 0.2 NDS. It effectively propagates multi-modal features while adaptively exploring their latent relationships and eliminating feature misalignment in the BEV domain. As reported in the last row, the combination of IB-Mamba and CB-Mamba yields the best performance, which benefits the subsequent query generation and feature decoding.

\noindent\textbf{Comparsion of different activation method} In PQG, we choose different methods to activate BEV features for extracting harder queries in the second stage. As reported in Table \ref{tab:query_generation}, utilizing a residual block yields a minor improvement. Our proposed HIA module further incorporates spatial cues from easy queries to facilitate the identification of remaining hard objects in driving scenarios and achieve higher accuracy gains. However, it is important to note that the cross-attention between queries and BEV features introduces additional computational overhead, which limits the number of queries and stages. We plan to address this issue in future work.

\noindent\textbf{Analysis of progressive decoder}
In the progressive decoder, both BEV-based and voxel-based decoders are employed for query refinement. As shown in Table \ref{query_decoder}, we explore various configurations of their layer counts (M and N) to analyze their impact on performance. In the first row, using only a single voxel-based decoder yields suboptimal accuracy, even underperforming the second row, which adopts a three-layer BEV decoder. This is because the BEV map at the end of the network contains richer contextual information and higher-level semantic features compared to voxels. Additionally, the BEV dense representation is more robust to point density variations than sparse voxel representation. Therefore, BEV features are critical for object detection in complex environments. However, as shown in the third row, adding a voxel decoder after the BEV decoder further improves performance by 0.2 mAP and 0.7 NDS, indicating that incorporating voxel features with fine geometric details can enhance regression accuracy of predicted boxes. However, further increasing M provides negligible performance gains. In Table \ref{query_decoder}, we demonstrate that the mixing block and self-attention can adaptively aggregate 3D shape features and long-range spatial information respectively, each contributing to performance improvements. As reported in the last row, combining them can achieve the best result.

\noindent\textbf{Analysis of on different scaling choices}
Following previous state-of-the-art methods, we explore increasing the model parameters or enlarging the input image size to achieve better accuracy. As indicated in Table \ref{scaling}, doubling the channels in the LiDAR backbone only provides a marginal improvement of 0.3 mAP and 0.2 NDS. In contrast, expanding the input image size from $256 \times 704$ to $384 \times 1056$ yields a boost of 0.9 mAP and 0.4 NDS. This demonstrates that providing higher-quality image voxels $V_I$ and BEV features $B_I$ is effective in enhancing the network's performance.

\subsection{Comparsion of Different Homogeneous Fusion Methods}
We further compare DDHFusion with other three representative homogeneous fusion methods mentioned in the introduction, including GraphBEV, DeepInteraction++, and UVTR, through extensive experiments.

First, we evaluate their performance under different weather and lighting conditions. Since DeepInteraction++ and UVTR do not provide relevant data in their papers, we reproduce their models using open-source codes to obtain their accuracy. As shown in Table~\ref{tab:weather}, DDHFusion achieves the highest mAP across all environments. Unlike UVTR, DDHFusion is not negatively affected by rainy conditions and performs better even in poor lighting environments. This robustness stems from the comprehensive intra-modal and cross-modal feature learning in DDHFusion, which progressively filters out corrupt information to learn robust latent representations. Additionally, as shown in Table~\ref{tab:dist}, we evaluate the performance of different models on object detection at varying distances, where ``Near,'' ``Middle,'' and ``Far'' represent ranges of $<20m$, $[20m, 30m]$, and $>30m$, respectively. Again, our proposed DDHFusion outperforms other methods across all distance ranges.
\begin{table}[tbp]
	\centering
	\renewcommand{\arraystretch}{1.2} 
	\caption{Performance comparisons under different lighting and weather conditions}
	\label{tab:weather}
	\begin{tabular}{c|c|c|c|c}
		\hline
		Method & Sunny & Rainy & Day & Night \\
		\hline
		\hline
		GraphBEV\cite{Song2024GraphBEVTR} & 70.1 & 70.2 & 69.7 & 45.1 \\
		Deepinteraction++\cite{Yang2024DeepInteractionMI} & 70.3 & 70.4 & 70.6 & 42.9 \\
		UVTR \cite{Li2022UnifyingVR} & 66.4 & 65.1 & 65.5 & 36.7 \\
		DDHFusion (ours) & \textbf{71.8} & \textbf{71.9} & \textbf{71.8} & \textbf{46.8} \\
		\hline
	\end{tabular}
\end{table}
\begin{table}[tbp]
	\centering
	\renewcommand{\arraystretch}{1.2} 
	\caption{Performance Comparisons with different ego distances}
	\label{tab:dist}
	\begin{tabular}{c|c|c|c}
		\hline
		Method & Near & Middle & Far \\
		\hline
		\hline
		GraphBEV\cite{Song2024GraphBEVTR} & 78.6 & 65.3 & 42.1 \\
		Deepinteraction++\cite{Yang2024DeepInteractionMI} & 80.0 & 66.5 & 42.4 \\
		UVTR \cite{Li2022UnifyingVR} & 77.2 & 61.2 & 37.0 \\
		DDHFusion (ours) & \textbf{82.3} & \textbf{67.8} & \textbf{44.2} \\
		\hline
	\end{tabular}
	
\end{table}
\begin{table}[htbp]
	\centering
	\renewcommand{\arraystretch}{1.2} 
	\caption{Comparison of memory cost and inference speed.
	}
	\label{tab:runtime}
	\begin{tabular}{c|c|c}
		\hline
		Method & Memory (MB) & FPS \\
		\hline
		\hline
		GraphBEV\cite{Song2024GraphBEVTR} & \textbf{9851} & \textbf{5.7} \\
		Deepinteraction++\cite{Yang2024DeepInteractionMI} & 13125 & 3.8 \\
		UVTR \cite{Li2022UnifyingVR} & 17640 & 3.6 \\
		DDHFusion (ours) & 12189 & 4.2 \\
		\hline
	\end{tabular}
\end{table}

\begin{figure*}[tbp]
\centering
\includegraphics[width=0.85\textwidth]{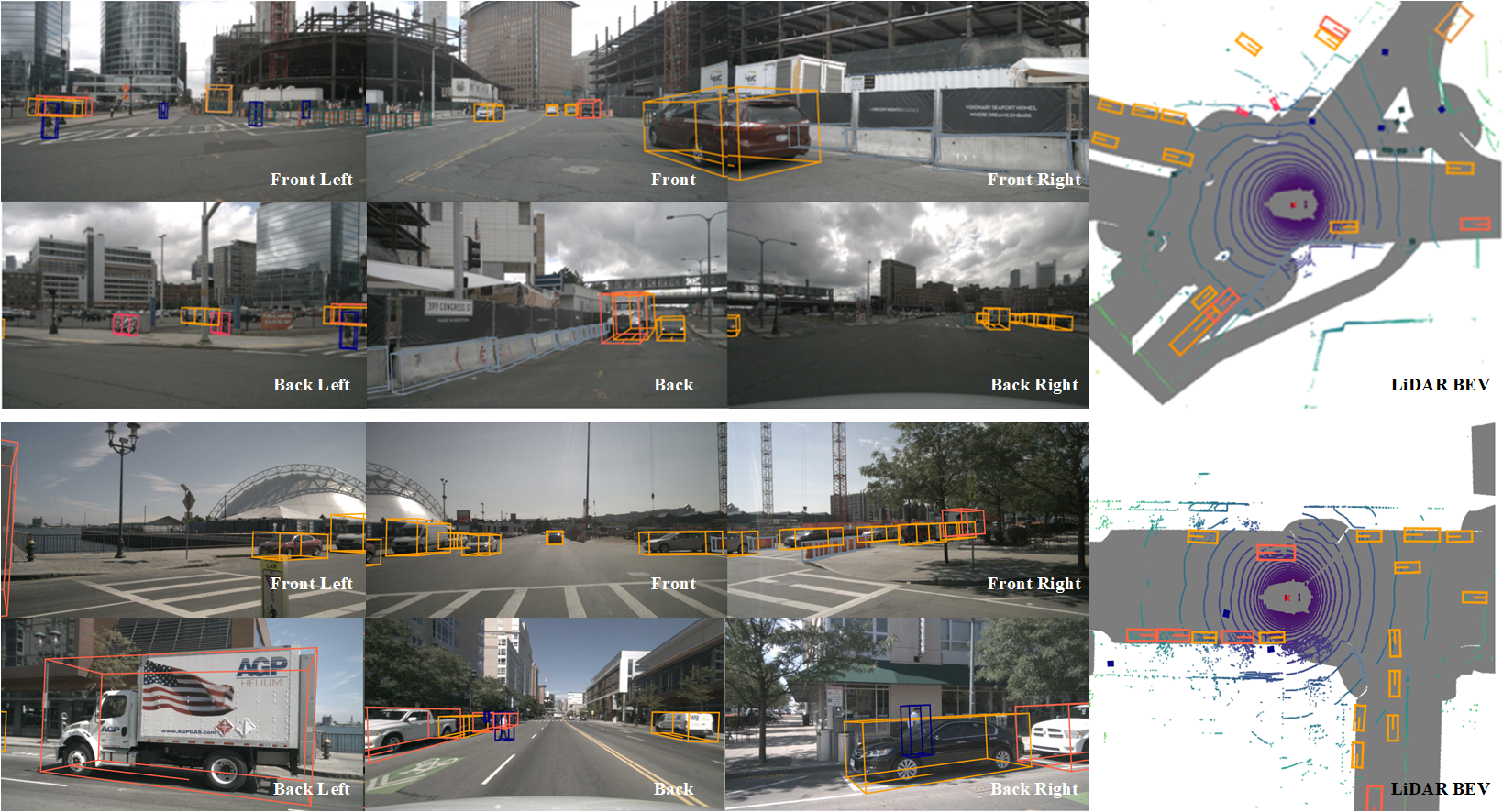} 
\caption{The visualization of DDHFusion on NuScenes validation set.} 
\label{fig:vis}
\end{figure*}
\begin{figure*}[htbp]
\centering
\includegraphics[width=0.85\textwidth]{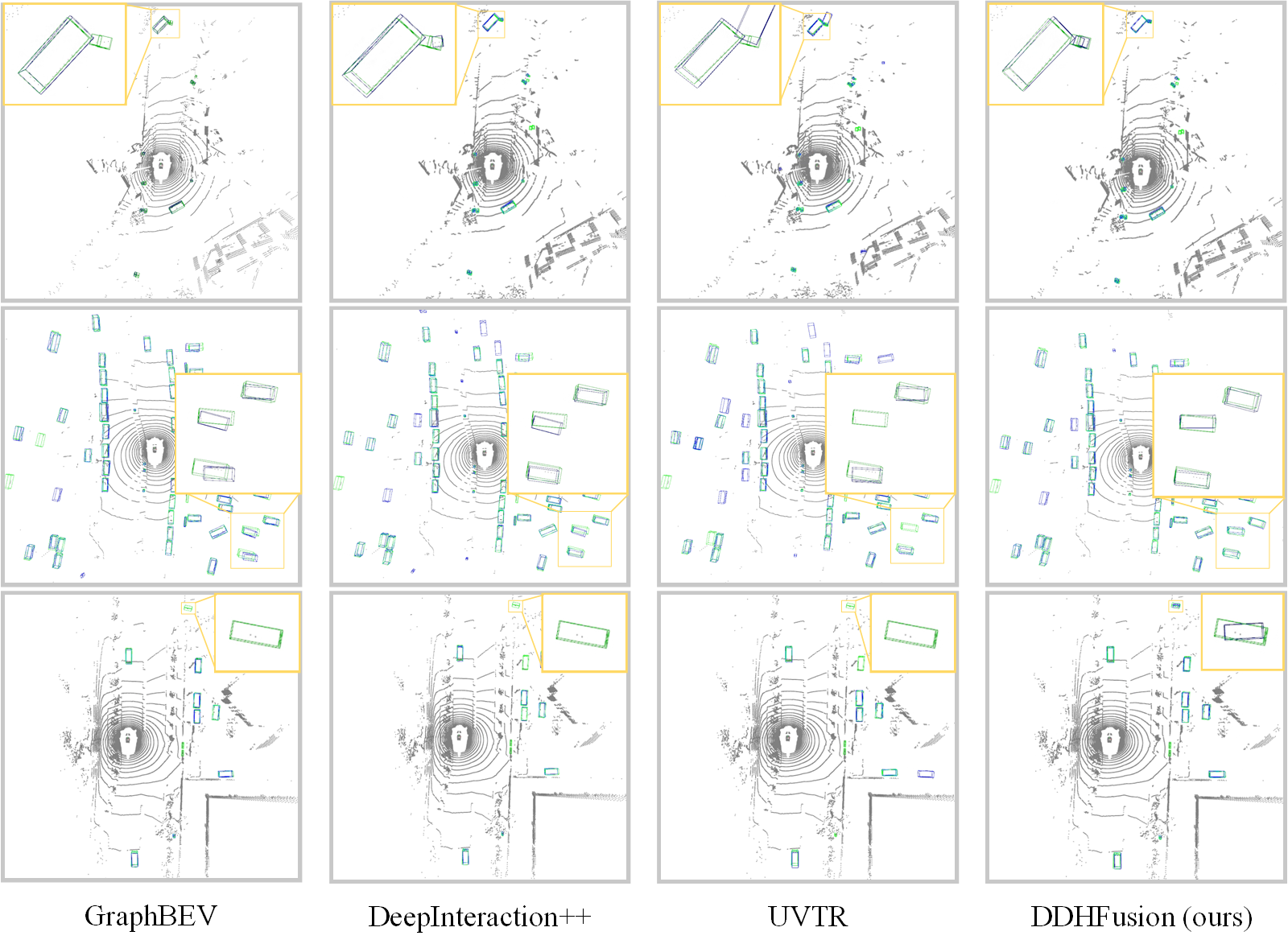} 
\caption{The visualization of different methods on NuScenes validation set.The \textcolor{green}{green} and \textcolor{blue}{blue} boxes represent groundtruths and inference results respectively} 
\label{fig:vis_comp}
\end{figure*}
Fig. \ref{fig:vis} showcases the detection results of DDHFusion, demonstrating its ability to accurately classify and localize objects of various categories in complex environments. Fig. \ref{fig:vis_comp} provides a qualitative comparison with other models. As shown in the first row, DDHFusion successfully detects three distant and crowded objects, demonstrating its ability to learn more discriminative representations from the two domains. In the second row, the bounding boxes generated by DDHFusion exhibit minimal deviations, which can be attributed to the rich structural details derived from the voxel domain. In the third row, it successfully identifies a small and distant object that other methods fail to detect, proving the effectiveness of the PQG module. These results highlight the superiority of our dual-domain homogeneous fusion framework.

As reported in Table \ref{tab:runtime}, we also compare their memory consumption and inference speeds. To ensure a fair comparison, all models are implemented within the mmdetection3D framework and tested on a single 4090D GPU. As shown in the last row, the computational resource consumption of DDHFusion is acceptable, which can be attributed to the efficiency of our Mamba-based fusion modules. GraphBEV is the most lightweight, as it performs fusion only in the BEV domain. In contrast, UVTR incurs the highest computational burden due to its retention of dense voxels, while DDHFusion mitigates this issue through the SAFS module. It is worth noting that DeepInteraction++ utilizes flash-attention for acceleration. Similarly, we believe that further hardware optimization of Mamba could lead to significant efficiency improvements.

\section{Conclusion}
In this work, we propose DDHFusion, which innovatively combines voxel and BEV homogeneous fusion paradigms and achieves exceptional performance. Another significant contribution is the successful extension of the recently popular Mamba architecture to the multi-modal 3D detection task, fully leveraging its versatility. Specifically, we employ cross-modal Mamba structures based on voxel or BEV representations in the HVF and HBF networks, which significantly enhance the receptive field during the fusion process and address the issue of feature misalignment. In the feature decoding stage, we also introduce PQG and PD to improve the target recall and regression accuracy of the network. In the future, we plan to explore the corresponding temporal fusion pipeline to further unleash the potential of DDHFusion.


%

\bibliographystyle{IEEEtran}
\bibliography{references}
\newpage

\end{document}